%% file: root.tex
\documentclass[IEEEtran, letterpaper, 10 pt, conference]{ieeeconf}

\IEEEoverridecommandlockouts

\input{preamble}

\begin{document}

% \makeatletter
% \apptocmd{\@maketitle}{\centering\insertfig}{}{}% insert the figure after authors
% \makeatother

% \maketitle
% \thispagestyle{empty}
% \pagestyle{empty}

\title{\LARGE \bf
\ours: Learning Agile Skills Switching for Humanoid Robots
% \author[1]{Yuen-Fui Lau$^*$}
% \author[1]{Qihan Zhao$^*$}
% \author[1]{Yinhuai Wang$^*$}
% \author[1]{Runyi Yu}
% \author[1]{Hok Wai Tsui}
% \author[1]{Qifeng Chen}
% \author[1]{Ping Tan}

% \affil[1]{The Hong Kong University of Science and Technology}

}

\author{Yuen-Fui Lau$^*$, Qihan Zhao$^*$, Yinhuai Wang$^*$, Runyi Yu, Hok Wai Tsui\\
Qifeng Chen$^\dagger$, Ping Tan$^\dagger$\\
\authorblockA{
The Hong Kong University of Science and Technology
}}

% Teaser image
\twocolumn[{%
\renewcommand\twocolumn[1][]{#1}%
\maketitle
\vspace{-0.8cm}
\thispagestyle{empty}
\pagestyle{empty}
\begin{center}
    \centering
    \captionsetup{type=figure}
     \includegraphics[width=1.0\textwidth]{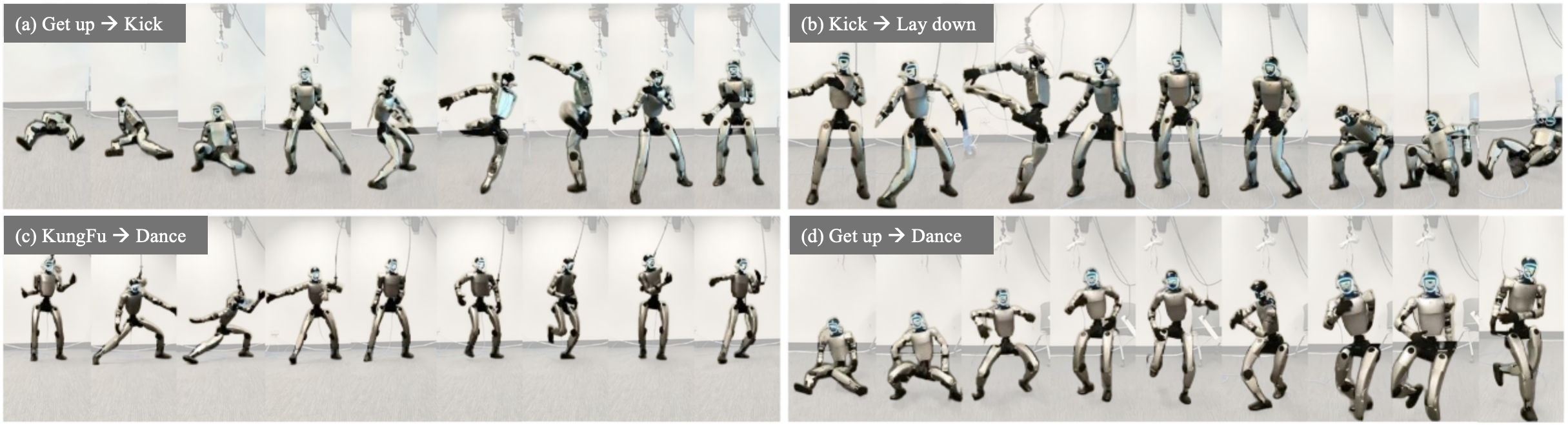}
    \vspace{2pt}
    \caption{We introduce \ours, a hierarchical whole-body control system that enables humanoid robots to perform agile and seamless skill switching between highly dynamic skills: (a) Get up to Kick; (b) Kick to Lay down; (c) Kungfu to Dance; (d) Get up to Dance.}
    \label{fig:teaser}
\end{center}
}]

\input{text/0_abstract}
\vspace{-0.05cm}
\blfootnote{$^*$ Equal contribution $^\dagger$Corresponding author}
\input{text/1_intro}
\vspace{-0.05cm}
\input{text/2_relatedwork}
\vspace{-0.05cm}
\input{text/3_method}

\vspace{-0.05cm}
\input{text/4_experiment}

\vspace{-0.05cm}

\input{text/5_conclusion}

% \bibliographystyle{IEEEtran} % Choose a style (plain, IEEEtran, apalike, etc.)
% \bibliography{egbib.bib}
\printbibliography

% \clearpage
% \input{text/6_appendix}

\end{document}

%% file: preamble.tex
\usepackage{amsmath}       % Advanced math typesetting
\usepackage{amssymb}       % Additional math symbols
\usepackage{amsfonts}      % Additional fonts for math
\usepackage{mathtools}     % Extensions to amsmath
\usepackage{bm}            % Bold math symbols
\usepackage{bbm}           % Black-board bold for numbers/sets
\usepackage{times}         % Times New Roman font
\usepackage{dsfont}        % For \mathds
\usepackage{xspace}        % Smart spacing after macros

\usepackage{graphicx}      % Include graphics
\usepackage{wrapfig}       % Wrap text around figures
\usepackage{float}         % Improved float control
\usepackage{stfloats}      % Allow two-column floats at bottom of page
\usepackage{capt-of}       % Captions outside float environment
\usepackage[font=small,labelfont=bf]{caption}      % Customize captions
\usepackage{subfigure}

\usepackage{booktabs}      % Better horizontal lines in tables
\usepackage{multirow}      % Merge multiple rows
\usepackage[flushleft]{threeparttable} % Table notes aligned left
\usepackage{arydshln}      % Dashed lines in tables

\usepackage[dvipsnames,table,xcdraw]{xcolor} % Extended color support

\definecolor{mydarkblue}{rgb}{0,0.08,0.45}
\definecolor{mydarkgreen}{RGB}{0, 139, 69}
\definecolor{mygreen2}{RGB}{0, 205, 0}
\definecolor{mybrown}{RGB}{139, 69, 19}
\definecolor{boxblue}{RGB}{79,173,234}
\definecolor{tablepeach}{RGB}{255, 240, 235}
\definecolor{tablepurple}{RGB}{248,235,252}
\definecolor{tableblue}{RGB}{235,241,255}

\usepackage{hyperref}                   % Smart cross-references
\usepackage[capitalise, nameinlink]{cleveref} % Smart referencing
\usepackage{xurl}                       % Line breaking for URLs
\definecolor{citecolor}{HTML}{D67B22}

\hypersetup{
    colorlinks=true,
    linkcolor=citecolor,
    urlcolor=citecolor,
    citecolor=citecolor,
}

\usepackage[linesnumbered,ruled,vlined]{algorithm2e} % Pseudocode

\usepackage{pifont}  % Symbols like \ding
\usepackage{dsfont}  % For double stroke numbers
\usepackage{lipsum}  % Dummy text
\usepackage{siunitx} % SI units formatting
\usepackage{multicol} % Multiple columns
\let\labelindent\relax
\usepackage{enumitem}

\usepackage[
    style=ieee,
    natbib=true,
    citestyle=numeric-comp,
    doi=false,
    isbn=false,
    url=false
]{biblatex} 
\usepackage{etoolbox} 
\AtBeginBibliography{\footnotesize} % Set bibliography font size
\newcommand{\ours}{{\color{citecolor}\textbf{Switch}}\xspace}

\newcommand\blfootnote[1]{%
  \begingroup
  \renewcommand\thefootnote{}\footnote{#1}%
  \addtocounter{footnote}{-1}%
  \endgroup
}

%% file: text/0_abstract.tex
\begin{abstract}
Recent advancements in whole-body control through deep reinforcement learning have enabled humanoid robots to achieve remarkable progress in real-world challenging locomotion skills. However, existing approaches often struggle with flexible transitions between distinct skills, creating safety concerns and practical limitations. To address this challenge, we introduce a hierarchical multi-skill system, \ours,  enabling seamless skill transitions at any moment.
Our approach comprises three key components: (1) a Skill Graph (SG) that establishes potential cross-skill transitions based on kinematic similarity within multi-skill motion data, (2) a whole-body tracking policy trained on this skill graph through deep reinforcement learning, and (3) an online skill scheduler to drive the tracking policy for robust skill execution and smooth transitions.
% Yinhuai Version
% We divide the SG according to skill semantics, with tracking targets selected along the corresponding graph path during skill execution. When transitioning between skills or when the robot significantly deviates from the current tracking target, our system performs online graph search to identify the optimal feasible path. This methodology enables efficient, stable, and real-time execution and transition of highly dynamic locomotion skills.
% Runyi Version
% We divide the SG based on skill semantics, with tracking targets selected along the graph edges during skill execution. 
For skill switching or significant tracking deviations, the scheduler performs online graph search to find the optimal feasible path, which ensures efficient, stable, and real-time execution of diverse locomotion skills.
Comprehensive experiments demonstrate that \ours empowers humanoid to execute agile skill transitions with high success rates while maintaining strong motion imitation performance.

% Tracking controllers driven by motion-capture (MoCap) reference trajectories have recently produced highly natural humanoid behaviors. However, free switching among heterogeneous skills remains difficult: the space of feasible entries between two sequences grows as $O(n^2)$ while the available data typically scales only as $O(n)$, leaving coverage gaps. In addition, open-loop advancement of the reference index renders controllers brittle under accumulated drift and external disturbances.

% We introduce TSS, a plug-and-play framework that jointly learns skills and their pairwise entry transitions in a single training stage. Concretely, we (i) construct a motion graph over all MoCap frames and add cross-skill edges based on pose-kinematic similarity; each cross edge carries a short transition buffer during which imitation is relaxed and smoothness/safety priors dominate, after which we resume strong frame-by-frame tracking of the target skill; and (ii) deploy an online reference search mechanism: whenever the system leaves the local region of attraction (RoA) of the current reference, we re-index to the nearest feasible reference frame—akin to motion matching but under closed-loop physics—thus recovering stable tracking without explicit transition trajectories.

% On a Unitree G1 humanoid, TSS relies only on non-privileged observations and a single-stage training regimen, yet achieves seamless, anytime switching among diverse skills (e.g., walking, get-up, lying down, striking/kicking maneuvers) and substantially improves the robustness of a tracking policy.

\end{abstract}

%% file: text/1_intro.tex
\section{Introduction}
%背景，引出论文主题
Recent advances in whole-body motion control, powered by simulation and reinforcement learning, have enabled humanoid robots to perform dynamic real-world skills such as highly agile acrobatics (e.g., flips and kicks) and expressive human-like dances\cite{truong2025beyondmimic, he2025asap, xie2025kungfubot, cheng2024expressive, ji2024exbody2}. For practical deployment, it is essential that robots not only perform individual skills robustly but also switch between them seamlessly. This paper aims to develop a unified framework for flexible skill switching in humanoid robots.

%现有方法的局限性
% Existing research predominantly focuses on learning individual skills in isolation. 

Multi-skill execution and switching typically require the existence of common states between skills, which imposes significant data requirements. Although using large-scale motion data to train a general whole-body tracking controller~\cite{chen2025gmt, yin2025unitracker, he2024omnih2o, li2025clone} allows skill switching via goal-conditioned tracking, this approach still relies on pre-defined trajectories with feasible transition states. Otherwise, it may lead to low switching success rate with unnatural movements such as tripping or stumbling \cite{helearning, cheng2024open, fu2024humanplus, he2024omnih2o, lu2025mobile}. Furthermore, due to the open-loop nature of tracking predefined motions, the controller is vulnerable to disturbances: once the tracking error exceeds a certain threshold, stability is compromised, potentially resulting in dangerous and uncontrolled behavior.

% Multi-skill execution and switching typically require the existence of common transition states between skills, which imposes significant data requirements and prevents switching at arbitrary moment. Although using large-scale motion data to train a general whole-body tracking controller~\cite{chen2025gmt, yin2025unitracker, he2024omnih2o, li2025clone} allows skill switching via reference tracking, this approach still relies on pre-defined trajectories with feasible transition states. Otherwise, it may lead to unnatural movements such as tripping or stumbling \cite{helearning, cheng2024open, fu2024humanplus, he2024omnih2o, lu2025mobile}. Furthermore, due to its open-loop nature in tracking predefined motions, the controller is vulnerable to disturbances: once the tracking error exceeds a certain threshold, stability is compromised, potentially resulting in dangerous and uncontrolled behavior. Traditional methods such as finite state machine (FSM)-based methods can achieve a certain level of skill switching, but they require manual definition of states and transition rules. This design inherently restricts systems to only one transition state at a time and struggles to cope with unanticipated scenarios. Furthermore, the number of states grows combinatorially with the increase of skills, making them difficult to scale and manage.

%我们的motivation
We observe that the main challenge in skill switching comes from poorly modeled transitions between different skills.
Since the controller lacks training on these transitions, the robot often fails when encountering such unseen states. One straightforward solution is to collect sufficient motion data that cover all possible inter-skill state-level transitions. However, this approach becomes prohibitively expensive as the number of skills grows, since the required transitions scale combinatorially.
%\ping{This paragraph needs to be improved to be more consistent with the termenologies in the Method section.}
To overcome this issue, we propose to establish the connections between motion states by exploring state similarity level across different skills, and use reinforcement learning to learn feasible transitions from such augmented data. Furthermore, instead of tracking a fixed reference trajectory in open-loop, the system should have online replanning capabilities. For example, when the tracking error exceeds a safety threshold, the system should replan a feasible trajectory to recover stable execution rather than allowing error to accumulate until failure.

%我们的方法介绍
% Based on these insights, we propose \ours, a hierarchical multi-skill controller as shwon in Fig~\ref{fig: pipeline}, that enables seamless skill switching at any moment as well as individual skill execution. Our framework consists of three key components: a graph-based data augmentation method, a unified whole-body tracking policy, and an online skill scheduler.
Based on these insights, we propose \ours, a hierarchical whole-body control system that enables seamless skill switching at any moment (Fig.~\ref{fig: pipeline}). Our framework consists of three key components: a skill graph as augmented dataset, a unified whole-body tracking policy, and an online skill scheduler.
%
% \ping{Add one sentence to briefly introduce your idea of inserting buffer states?}
% Specifically, we treat motion states as nodes in a configuration space and transitions cost between frames as edges to form a graph. Given multiple sequences of different skills, we connect some unlinked nodes based on state similarity level, thereby automatically creating opportunities to learn state transition between skills. This process significantly enriches the reference motion dataset.
%
Specifically, we treat motion data frames as graph nodes and transitions between frames as directed edges. Given multiple skill sequences, we insert edges between previously disconnected nodes based on similarity, thus automatically creating transition opportunities. For newly added transitions with significant gaps, we insert buffer nodes to enhance switching smoothness and stability. The number of buffer nodes is calculated by the distance between the endpoints of each edge, with larger gaps leading to more buffer nodes. This process significantly enriches and augments the reference motion dataset.
We then train a unified tracking policy in simulation to control the robot for executing both the skills and the learned transitions. A buffer-aware imitation mechanism is introduced to train the policy by reinforcement learning with transitions involving these buffer nodes. 
Finally, an online skill scheduler provides real-time tracking guidance to the policy. When a skill switch is intended or when tracking failure is detected, the scheduler performs real-time motion planning via shortest-path search over the skill graph. This generates new tracking targets to guide the tracking policy toward smooth skill switching or recovery.
The proposed \ours is designed to be versatile across a wide range of locomotion skills, significantly improving both skill transition success rate and execution stability compared to existing methods. Experimental results on the Unitree G1 demonstrate that our system can be successfully deployed in the real world to achieve continuous transitions between dynamic and complex motor skills such as Kung Fu squats, back kicks, and rapid recoveries.

%% file: text/2_relatedwork.tex
\section{Related Work}

\subsection{Learning-Based Motion Tracking}
Learning-based tracking has achieved significant progress in character animation, enabling lifelike whole-body motions by imitating individual references~\cite{peng2018deepmimic, wang2023physhoi, xu2025parc, wang2025skillmimic} or learning universal tracking~\cite{luo2023perpetual, luo2024universal, xu2025intermimic, tessler2024maskedmimic, tessler2025maskedmanipulator}. Recent efforts extend these techniques to real-world robots, focusing on improving accuracy~\cite{he2025asap, xie2025kungfubot, ji2024exbody2, ze2025twist, li2025clone}, generalization~\cite{yin2025unitracker, shao2025langwbc, chen2025gmt, wang2025experts, xue2025leverb, wang2026humanx}, and robustness~\cite{truong2025beyondmimic, ji2024exbody2, li2025hold}. For accuracy, some works~\cite{ji2024exbody2, ze2025twist} achieve low short-term tracking errors, while CLONE~\cite{li2025clone} mitigates long-horizon drift by incorporating an LiDAR odometry. UniTracker~\cite{yin2025unitracker} further improves tracking precision by introducing a fast adaptation module. 
For generalization, GMT~\cite{chen2025gmt} employs MoE to handle diverse motion tracking, while UniTracker leverages a teacher-student framework to enable scalable tracking across thousands of motions.
For robustness, SoFTA~\cite{li2025hold} proposes to use a slow-fast agent design for end-effector control. BeyondMimic~\cite{truong2025beyondmimic} enhances robustness by distilling a diffusion model.

Despite these advancements, inter-sequence transitions remain a challenge, often resulting in inefficiencies or failures during skill switching. To bridge this gap, we introduce short buffer states between pre-linked sequences and propose a method to explicitly learn these transitions.

\subsection{Skill Switching and Sequencing}
\label{sec:rw-switch}

Skill switching and sequencing have been investigated across animation and robotics through kinematic composition, attractor-based primitives, hierarchical decision layers, and reachability-aware planning.
Early work in character animation~\cite{Kovar2002MotionGraphs,Heck2007ParametricMotionGraphs,Zhao2008GoodConnectivity,Beaudoin2008MotionMotifGraphs,Reitsma2007EvaluatingMG,Kovar2004AutomatedExtraction} addresses switching by organizing motion data into discrete graphs and local blends.
Motion Graphs model a dataset as a directed graph whose nodes are poses or short clips and whose edges are kinematically compatible transitions; long behaviors arise by walking paths on the graph~\cite{Kovar2002MotionGraphs}. Parametric Motion Graphs and runtime blending further improve connectivity by interpolating families of motions~\cite{heck2007parametric}. While these techniques achieve visually smooth transitions (e.g., via LERP/SLERP interpolation), they operate primarily at the kinematic level and do not guarantee dynamic feasibility or stability on physical robots, revealing a persistent kinematics–dynamics gap. 

In robotics, reusable skills are often encoded as dynamical systems with attractors, as in DMP/SEDS, enabling time/space retargeting and composition with stability guarantees~\cite{ijspeert2013dynamical,khansari2011learning}. Geometric formulations such as RMPflow provide principled rules to combine task policies while preserving stability properties~\cite{cheng2018rmp}. Orthogonally, hierarchical RL abstracts time by letting a high-level policy select among low-level options or subgoals~\cite{sutton1999between}, and a large body of work discovers/composes skills (e.g., skill chaining and skill graphs) to handle long-horizon tasks~\cite{konidaris2009skill,bagaria2021skill}. Bridging selection and planning, PRM-RL admits an edge only if a local controller can reliably traverse it, reducing long-range behavior to graph search over feasible edges~\cite{faust2018prm}. In control, Reference/Command Governors and tube-based MPC adjust or filter targets so that tracking remains within constraints, while viability/capture-basin concepts formalize the set of states from which safe progress is possible~\cite{bemporad2002reference,rakovic2005simple,aubin2011viability}.

Building upon these principles, our approach, \ours, frames switching as \emph{state-conditioned reference selection}. Specifically, it uses a short-edge skill graph to provide the tracker with feasible subgoals that lead to the target skill. By combining low-level learned tracking with high-level graph search, \ours enables smooth transitions and robust recoveries, even with limited transition data or poorly timed user commands.

%% file: text/3_method.tex
\section{Method}

\begin{figure*}[t!]
\begin{center}
  \includegraphics[width=\textwidth]{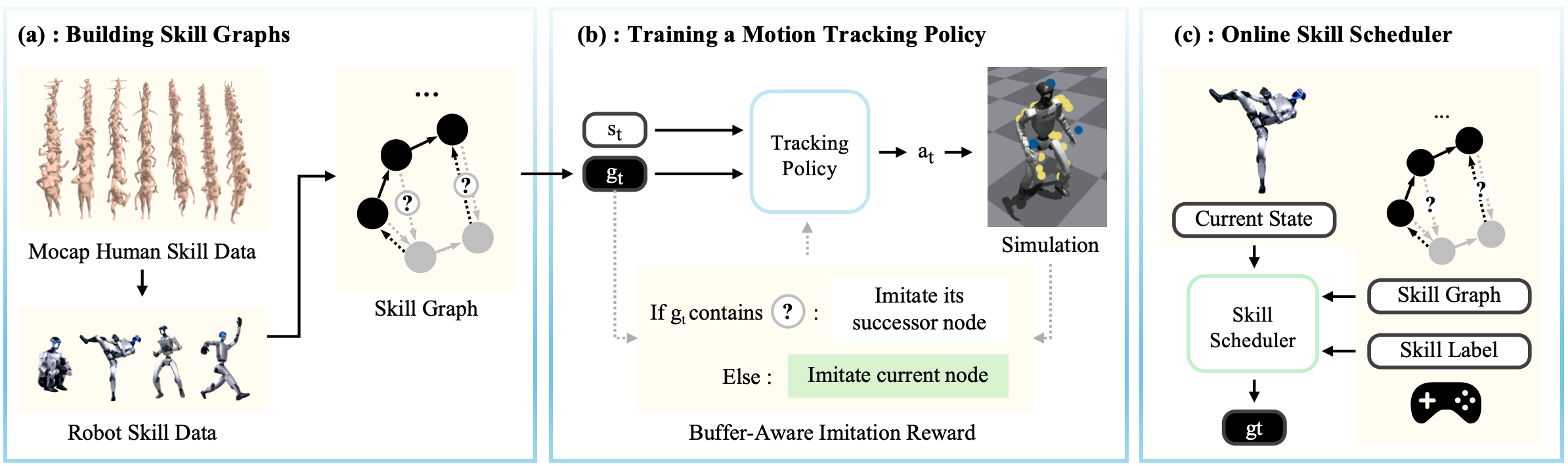}
\end{center}
  \caption{ The \ours system: \textbf{(a)} We retarget human motion capture skills onto the robot. We then construct a skill graph where frames serve as nodes (solid circles) and temporal transitions as edges (solid arrows). \textbf{black} and \textcolor{gray}{\textbf{gray}} represents two different skills. Based on frame similarity, we add additional transitions between different skills (dashed arrows). When the added transition has significant gap, we insert buffer nodes (question marks) to enhance switching feasibility. \textbf{(b)} The assembled Skill Graph is used for training a motion tracking policy. During normal operation, imitation rewards measure the distance between simulated results and reference frame. When referencing a buffer node, we measure with its
first non‑buffer successor node. \textbf{(c)} During deployment, users can change skill labels at any time, and our online scheduler plans optimal paths through the skill graph based on the robot's current state, providing real-time targets to guide the tracking policy.}
    \vspace{-0.6cm}
  \label{fig: pipeline}
\end{figure*}

\subsection{Building a Skill Graph for Data Augmentation} 

% Training a universal whole-body motion tracker to enable state transitions between different skills is challenging, as it requires massive, often unavailable high-quality motion data. Alternatively, stitching motion trajectories between all pairs of states from different skills leads to convergence issues, especially since transitions between distant state neighborhoods may be physically infeasible. To address these problems, we propose the Skill Graph (SG) as shown in fig.~\ref{fig: pipeline}(a), a directed graph that models feasible state transitions: its nodes \(V\) are reference states from multiple skills, and its edges \(E\) link distinct nodes representing similar states across different skills, with each edge storing an inter-state transition cost. This cost is calculated by a similarity function \(\psi\) that compares the joint positions, joint velocities, and local rigid body positions of two states. For a state \(\hat{s}_i^{\phi_p}\) from skill \(\phi_p\), we identify its most similar state \(\hat{s}_j^{\phi_q}\) from skill \(\phi_q\) via \(\hat{s}_j^{\phi_q} = arg max_{s \in \phi_q} \psi(\hat{s}_i^{\phi_p}, s)\)) and establish a connection between them, constructing an augmented motion trajectory \(\{\hat{s}_0^{\phi_p}, ..., \hat{s}_i^{\phi_p}, \hat{s}_j^{\phi_q}, ..., \hat{s}_L^{\phi_q}\}\) to enrich the reference motion data.

Training a universal whole-body motion tracker to enable state transitions between different skills is challenging, as it requires massive, often unavailable high-quality motion data. Alternatively, stitching motion trajectories between all pairs of states from different skills leads to convergence issues, especially since transitions between distant state neighborhoods may be physically infeasible.  To address these issues, we need to establish a feasible state mapping for each neighborhood as shown in fig.~\ref{fig: pipeline}(a) and identify physically achievable state for skill switching.

% To achieve state transitions between two skills, one straightforward way is to train a universal motion tracker capable of generalizing to all possible state transitions. However, learning such a general whole-body tracking controller requires an enormous amount of high-quality, diverse motion data that are costly to acquire and often unavailable. Given a limited amount of motion data, another direct approach to learn these transitions involves stitching motion trajectories from any pair of states from skills $\phi_{p}$ and $\phi_{q}$, and learning the transition between every combination of reference states $(\hat{s}^{\phi_{p}}_{i}, \hat{s}^{\phi_{q}}_{j})$, where $\hat{s}^{\phi_{p}}_{i} \in \phi_{p}$ and $\hat{s}^{\phi_{q}}_{j} \in \phi_{q}$ are two states from these different skills. However, this approach raises convergence issues, both in theory and in practice. For example, when neighborhoods are far apart, transitions between state pairs associated with these distant neighborhoods may be physically impossible. To address these issues, we need to establish a feasible state mapping for each neighborhood as shown in fig.~\ref{fig: pipeline} (a) and identify physically achievable state for skill transitions. 

% % Filtering out transitions that are not physically achievable ensures convergence.

% % \ping{what is the difference between the first and this second approach if you will stitch any pair of states?}
\subsubsection{Definition of Skill Graph}
\label{sec:sg-def}
We represent candidate state-to-state transitions with a directed, weighted graph
$\mathcal{G}=(V,E,w)$, with edge weight function $w:E\!\to\!\mathbb{R}_{\ge 0}$.
Let $\Phi=\{\phi_1,\dots,\phi_K\}$ be the set of skills. Each skill
$\phi\in\Phi$ provides a reference sequence $\{\hat{s}^{\phi}_t\}_{t=1}^{T_\phi}$,
where $T_\phi$ is its length. We collect all reference states as
\begin{equation}
V \;=\; \bigcup_{\phi\in\Phi}\big\{\hat{s}^{\phi}_t\big\}_{t=1}^{T_\phi}.
\end{equation}
Edges $E$ consist of (a) all consecutive pairs within each reference sequence
and (b) cross-skill connections between sufficiently similar states.
The meaning of the edge weight $w(u,v)$ is phase-dependent and will be
instantiated in Sec.~\ref{sec:sg-build} (training-time) and
Sec.~\ref{sec:weight_deployment} (deployment-time).

\subsubsection{Construction of Skill Graph}
\label{sec:sg-build}
\paragraph{Training-time edge weight (distance)}
During graph construction we instantiate the edge weight primarily via a state distance.
In the local frame (with global $x$–$y$ translation and yaw/twist removed), define
\begin{equation}
  d\!\left(\hat{s}^{m},\hat{s}^{n}\right)
  \;=\;
  \big\|q^{m}-q^{n}\big\|_{1}
  + \big\|\dot q^{m}-\dot q^{n}\big\|_{1}
  + \big\|\hat p^{m}-\hat p^{n}\big\|_{1}.  
\end{equation}
We then assign the training-time edge weight as
\begin{equation}
w_{\text{train}}(u,v) \;=\;
\begin{cases}
1, & \text{if } (u,v) = (\hat{s}^{\phi}_t,\hat{s}^{\phi}_{t+1}) \\[2pt]
d(u,v), & \text{otherwise.}
\end{cases}
\end{equation}
% \paragraph{Training-time edge weight (distance)}
% During graph construction we instantiate the edge weight via a state distance.
% In the local frame (with global $x$–$y$ translation and yaw/twist removed), define
% \begin{equation}
%   d\!\left(\hat{s}^{m},\hat{s}^{n}\right)
%   \;=\;
%   \big\|q^{m}-q^{n}\big\|_{1}
%   + \big\|\dot q^{m}-\dot q^{n}\big\|_{1}
%   + \big\|\hat p^{m}-\hat p^{n}\big\|_{1}.  
% \end{equation}
% We set the training-time edge weight as
% \begin{equation}
% w_{\text{train}}(u,v) \;=\; d(u,v).    
% \end{equation}

\paragraph{Cross-skill connections}
Sample a source skill $\phi_p\in\Phi$ and index $i\in\{1,\dots,T_{\phi_p}\}$,
then connect $\hat{s}^{\phi_p}_i$ to its nearest neighbor in a different target
skill $\phi_q\in\Phi$ ($\phi_q\neq\phi_p$) under $d$:
\begin{equation}
j = \arg\min_{t\in [T_{\phi_q}]} d(\hat{s}^{\phi_p}_i,\hat{s}^{\phi_q}_t),
\qquad
E \leftarrow E \cup \{(\hat{s}^{\phi_p}_i,\hat{s}^{\phi_q}_j)\}.
\end{equation}

\subsection{Training a Whole-Body Tracking Controller in Simulation}

\subsubsection{Problem Formulation} 

We formalize the multi-skill learning problem as a Markov Decision Process (MDP) $M=\langle S,A,P,r,\gamma\rangle$, where $s\in S$ is the robot state, $a\in A$ is the action, $P(\cdot\mid s,a)$ is the state-transition kernel over $S$, $r$ is the reward function (covering motion imitation and regularization terms), and $\gamma\in[0,1)$ is the discount factor. At time $t$, the policy input is $o_t=\langle s^o_t, g_t\rangle$, where $s^o_t\triangleq\langle a_{t:t-n}, f_{t:t-n}, q_{t:t-n}, \dot q_{t:t-n}, \omega^{\mathrm{root}}_{t:t-n}\rangle$ with $a_t\in\mathbb{R}^{23}$ the previous actions (target joint positions), $f_t\in\mathbb{R}^3$ the body-frame gravity vector, $q_t,\dot q_t\in\mathbb{R}^{23}$ the joint positions/velocities, and $\omega^{\mathrm{root}}\in\mathbb{R}^3$ the root angular velocity. The guidance input is $g_t=\langle\hat s^g_t,\kappa_t\rangle$, where $\hat s^g_t={\hat p^g_t}$ encodes reference local rigid-body positions with $\hat p^g_t\in\mathbb{R}^{3\times 23}$, and $\kappa_t\in\mathbb{N}0$ is an integer indicating the remaining number of steps to the end of a buffer segment (set $\kappa_t=0$ if the target is not a buffer node). The instantaneous reward is $r_t=r(s^o_t,\hat s^g_t)$, and the action $a_t\in\mathbb{R}^{23}$ specifies desired joint positions tracked by a PD controller. Our objective is to learn a robust policy $\pi$ that enables smooth skill transitions while maintaining high-fidelity motion imitation; given reference sequences $\phi_p=\{\hat s^p_0,\dots,\hat s^p_{T_{\phi_p}}\}$ and $\phi_q=\{\hat s^q_0,\dots,\hat s^q_{T_{\phi_q}}\}$ from any two skills, we maximize $\mathbb{E}[ ~\sum_{t=0}^{T-1} \gamma^{t-1} r_t ~]$. We adopt Proximal Policy Optimization (PPO) for training.

\subsubsection{Reference State Inititaion}

Proper task initialization is critical for reinforcement learning (RL) training, and previous works have employed Reference State Initialization (RSI)~\cite{peng2018deepmimic,haarnoja2024learning,he2025hover,peng2022ase,he2025asap} to enhance the learning process. However, directly applying RSI to our augmented motion trajectories hinders the learning of skill transitions when states are sampled uniformly across the entire motion sequence, some sampled states occur after skill transitions, leading to scenarios where the agent fails to experience skill transitions when initialized from these states. To resolve this issue, we modify RSI by only sampling initial states that are \(n\) steps before skill transitions. This adjustment ensures that every sampled initial state allows the agent to encounter a skill transition during training.

\subsubsection{Buffer-aware Imitation Learning}

Basic imitation learning methods are constrained by their reliance on explicitly defined reference states for supervision, yet cross-skill transition states are often unavailable. Furthermore, their quantity grows polynomially with the number of skills which make data collection prohibitively expensive and their sparsity may render single-step transitions infeasible. To address this, we insert \(N\) buffer nodes between states, with the number \(N\) determined by the similarity level between the states; these nodes act as temporal buffers to bridge distant transitions, enabling the agent to explore viable paths instead of relying on unavailable predefined reference trajectories. Formally, the motion trajectory is constructed as:
\begin{equation}
\left\{\hat{s}_{0}^{\phi_{p}}, \ldots, \hat{s}_{i}^{\phi_{p}}, \underbrace{s_{\emptyset}, \ldots, s_{\emptyset}}_{N}, \hat{s}_{j}^{\phi_{q}}, \ldots, \hat{s}_{L}^{\phi_{q}}\right\},
\end{equation}

where \(N\) is computed similarly to~\cite{yu2025skillmimic}. Unlike~\cite{yu2025skillmimic}, which omits reward computation during the buffer stage, we use the target state \(\hat{s}_{j}^{\phi_{q}}\) to calculate rewards in this phase as demonstrated in fig.~\ref{fig: pipeline}(b). This guides the model toward the target state, facilitating convergence and preventing drastic deviations that could cause performance collapse.

\subsubsection{Unified Imitation Rewards}

We observe that standard imitation rewards underweight high-frequency foot–ground events (e.g., dancing, martial arts), causing the policy to act conservatively and degrade agility and motion fidelity. To explicitly supervise contact events, we add a Foot–Ground Contact Reward (FGR).

Let $G$ denote the index set of foot end-effectors that can contact the ground (e.g., $G=\{\mathrm{LF},\mathrm{RF}\}$). For each $i\in G$ and time $t$, let $c^{\mathrm{ref}}_{i,t}\in\{0,1\}$ be the reference contact label (obtained via self-labeling adapted from~\cite{xie2025kungfubot}) and $c_{i,t}\in\{0,1\}$ be the measured contact status. We define
\begin{equation}
r^{\mathrm{FGR}}_{t}
\;=\;
\exp\!\Big(-\,\lambda_{\!c}\,\sum_{i\in G}\,|c_{i,t}-c^{\mathrm{ref}}_{i,t}|\Big),    
\end{equation}

The overall reward function \(r_t\) is the sum of three components: task rewards for achieving precise whole-body tracking, penalties for preventing undesirable motions, and a regularizations for refining motion.

\subsubsection{Skills Training Curriculum}

\begin{table}[t!]
\centering
\caption{Reward Specification}
\renewcommand{\arraystretch}{1.5}
\resizebox{0.49\textwidth}{!}{
\label{tab:pretrain_reward}
\begin{tabular}{lr|lr}
% \hline
\toprule[0.8pt]
\multicolumn{4}{c}{\textbf{Task Reward}}  \\ 
\midrule[0.8pt]
Body position                         & $1.125$      & VR 3-point                             & $1.8$      \\ 
Body position (feet)                  & $2.3625$      & Body rotation                          & $0.5$      \\ 
Body angular velocity                 & $0.5$      & Body velocity                          & $0.5$      \\ 
DoF position                          & $0.75$     & DoF velocity                           & $0.5$      \\ 
FGR                                   & $1.8$      & & \\ 
\midrule[0.8pt]
\multicolumn{4}{c}{\textbf{Penalty}}  \\ 
\midrule[0.8pt]
DoF position limits                   & $-10.0$    & DoF velocity limits                   & $-5.0$     \\ 
Torque limits                         & $-5.0$     & Termination                            & $-200.0$   \\ 
\midrule[0.8pt]
\multicolumn{4}{c}{\textbf{Regularization}}   \\ 
\midrule[0.8pt]
Torques                               & $-1 \times 10^{-5}$ & Action rate                            & $-0.5$                                \\ 
\bottomrule[0.8pt]
\end{tabular}}
\end{table}

% Training a policy to master agile skill transitions in simulation is  challenging, especially in the early stage with no prior knowledge of locomotive skills, the agent even struggles to track a single skill.

Training a policy to achieve smooth skill transitions in simulation remains challenging, particularly during early stages when the agent lacks prior knowledge of locomotor skills.
To address this, we introduce three progressive training curricula for step-by-step learning, ensuring that the agent builds foundational locomotive skills first before tackling complex skill transitions. 

\begin{itemize}
    
\item \textit{Skill Augmentation Curriculum} initializes with a 10\% augmented motion
trajectories and 90\% of trajectories are single-skill to let the agent focus on learning basic locomotion through consistent single-skill practice. As training progress, the probability is incrementally raised to 50\%, systematically increasing demands for exploring and mastering skill transitions. 

\item \textit{Reward Penalty Curriculum}~\cite{he2025asap, yin2025unitracker} applies to regularization and penalty terms: it allows the policy to prioritize core tasks (e.g., basic locomotion or single-skill tracking) initially without being overwhelmed by extra constraints. By introducing these components incrementally, the policy can develop more reasonable, well-adjusted behaviors over time, rather than struggling to balance multiple objectives from the start

\item \textit{Termination Curriculum}~\cite{he2025asap}  terminates episodes when the robot deviates from the reference motion beyond a threshold: starting with a generous threshold, it is gradually tightened to incrementally raise tracking demands, facilitating the learning of agile motions and improving tracking performance.
\end{itemize}

\subsection{Deploy: Online Scheduling with Search-Based Planners}

\begin{figure*}[t!]
\centering
\includegraphics[width=\linewidth]{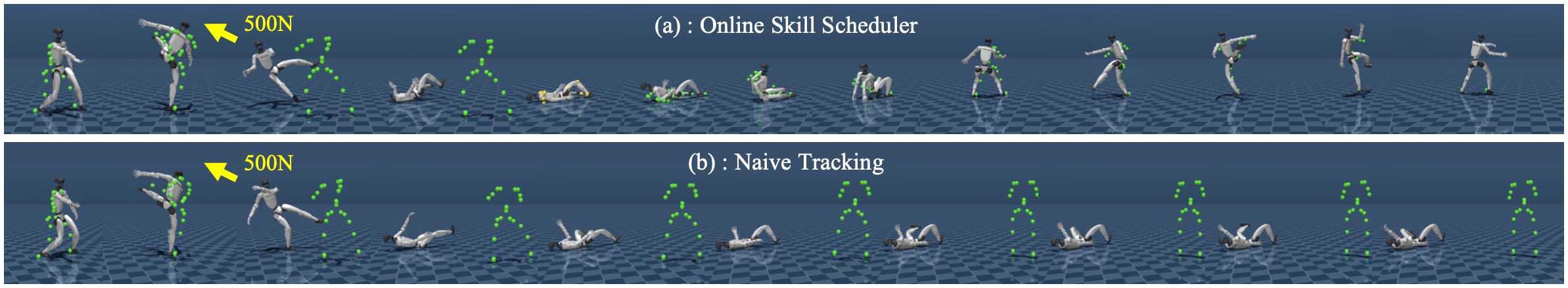}
\vspace{2pt}
\caption{Effect of the online skill scheduler. (a) When the robot performs a kicking skill, we apply a 500N disturbance force causing it to fall. This triggers the scheduler to automatically replan based on tracking errors, selecting an appropriate path (utilizing get-up skill segments) to resume the kicking execution. (b) Without the scheduler, tracking errors will gradually amplify and unable to recover.}
\vspace{-0.6cm}
\label{fig:scheduler_disturb}
\end{figure*}

After training, the tracking policy demonstrates seamless tracking of skills and skill switching, while maintaining robustness against certain level of disturbances. Nevertheless, a naive approach that merely follows a predefined motion reference suffers from active runtime controllability~\cite{chen2025gmt,he2025asap}. This deficiency becomes particularly problematic when addressing accumulated tracking errors or severe perturbations, as illustrated in Fig.~\ref{fig:scheduler_disturb}b. We therefore employ an \emph{online skill scheduler} in deployment for switching and safety. When a trigger occurs, it invokes a \emph{planner} to produce an entry and a path to a target set $T$, from which we synthesize the reference for the controller.

\subsubsection{Shared Graph Model and Planner Interface}\label{sec:weight_deployment}

In deployment, we reuse the identical skill graph \(\mathcal{G}\) constructed during training. Reference frames serve as nodes, and directed edges correspond to the transitions established in training. Given that collecting or learning transitions for all \(O(n^2)\) skill pairs is impractical, this graph acts as a data-driven prior—restricting the planner to search only over edges sampled or synthesized during training. 

Each edge is assigned a nonnegative deployment-time weight
$w_{\text{deploy}}(u \!\to\! v) = c{(u \!\to\! v)}$. We set $c=1$ for consecutive edges $(\hat{s}^{\phi}t \!\to\! \hat{s}^{\phi}{t+1})$ within the same skill.
For cross-skill edges $(\hat{s}^{\phi_p}_i \!\to\! \hat{s}^{\phi_q}_j,\;\phi_p\neq\phi_q)$
we use a distance-plus-penalty form:
\begin{equation}
  c(u \!\to\! v)
  \;=\; d(u,v)
  \;+\; \lambda_{\mathrm{sw}}\,[\!\![\,\text{skill}(u)\neq \text{skill}(v)\,]\!\!],    
\end{equation}
where $d(\cdot,\cdot)$ is the state distance defined in Sec.~\ref{sec:sg-build}
% , and $\mathrm{other}(\cdot,\cdot)$ denotes optional terms (e.g., RoA violations or
% latency)
. The total path weight is the sum of edge weights, and planning is
posed as a shortest-path-to-set problem on $\mathcal{G}$:
\begin{equation}
\min_{\pi: v \to T} \sum_{e \in \pi} c(e),    
\end{equation}

% which is equivalent to defining $V(t)=0$ for $t\in T$ and
% \[
%   V(v) \;=\; \min_{(v \to u)\in E}\big[\,c(u \!\to\! v) + V(u)\,\big],
% \]
% with the shortest path reconstructed via a next-hop map.

% We also expose a planner-agnostic interface:
% \[
% \text{Plan}(\mathcal{G}, T, x, \text{candidates}) \to (\text{entry}, \text{path}),
% \]
% where \(x\) denotes the robot’s current state, and \(\text{candidates}\) is a small entry set selected by similarity in the local pose–velocity subspace.

\subsubsection{Planner Choices}\label{sec:planner_choices}
\paragraph{Graph-Search planner}
We run a reverse multi-source shortest path from $T$ to obtain a value function $V^{\*}$ and a next-hop map $\texttt{NextHop}$. Given an entry, the path is reconstructed by iterating $\texttt{NextHop}$. This global view can leverage overlaps between RoAs e.g., briefly proceeding in the current skill to a nearby feasible state before merging into the target (see Fig.~\ref{fig:scheduler_disturb}).

\paragraph{Nearest-Neighbor (NN) planner}
We select the entry by nearest-neighbor similarity (optionally among a small candidate set) and form a single-hop or short-hop transition toward $T$ without global graph search. This planner has minimal latency and a simple implementation; in our hardware deployment we use the NN variant.

\subsubsection{Problem A: Intent-Driven Switching}
Given a commanded skill, we define the target set as its prefix $T_{\text{cmd}}$ (e.g., the first $\tau$ fraction of frames) and seek a plan that reaches any $t\in T_{\text{cmd}}$ with minimum cost.

\emph{Entry check and selection.}
We compute similarity in the local pose–velocity subspace. If $\mathrm{sim}\!\le\!A$, the state lies in a reference RoA and we attach directly. If $\mathrm{sim}\!\ge\!B$, we defer switching and enter emergency stop (e-stop). For $A<\mathrm{sim}<B$, we evaluate the top-$k$ candidates by a composite score
\begin{equation}
J(v)=\lambda_{\text{cost}}\,c(x,v)+\underbrace{V^{\*}(v)}_{\text{if available}},
\end{equation}
where $V^{\*}$ is used when the Graph-Search planner is selected; with NN this term is omitted.
The scheduler then calls \texttt{Plan}$(\mathcal{G},T_{\text{cmd}},x,\text{candidates})$ and installs the synthesized reference.

\subsubsection{Problem B: Safety Recovery (emergency stop)}\label{sec:estop_recovery}

Safety recovery is triggered when safety checks detect excessive divergence either when the similarity \(sim \geq B\) during tracking, or when the best candidate still violates \(B\) during entry selection. By default, recovery retains the original commanded target set \(T_{cmd}\). The planner is allowed to route through recovery skills as intermediate segments on the path to \(T_{cmd}\) whenever this yields a lower-cost or safer plan. For the Graph-Search planner, we set \(T = T_{cmd}\) and compute the shortest path, with recovery skills naturally integrated into the path. For the Nearest-Neighbor (NN) planner, if a direct jump toward \(T_{cmd}\) is unsafe, we optionally introduce a recovery target \(T_{rec}\)  and adopt a two-stage plan: first reach \(T_{rec}\), then re-plan to \(T_{cmd}\). During e-stop, we override the policy with a damping controller and wait until the system is stationary (e.g. angular velocity of the root below thresholds) before executing the recovery plan.

% % \emph{Triggers and formulation.}
% % We enter emergency stop (e-stop) when the safety check indicates excessive divergence: either during tracking when $\mathrm{sim}\!\ge\!B$, or during entry selection when the best candidate also violates $B$. 
% % By default, recovery keeps the \emph{same} commanded target set $T_{\mathrm{cmd}}$: the planner is free to route through recovery skills (e.g., \emph{get-up}) as intermediate segments on the path to $T_{\mathrm{cmd}}$ whenever this yields a lower-cost or safer plan. 
% % \emph{With the Graph-Search planner}, we thus set $T=T_{\mathrm{cmd}}$ and compute the shortest path. Recovery skills appear naturally as part of that path. 
% % \emph{With the NN planner}, if a direct jump toward $T_{\mathrm{cmd}}$ is unsafe, we optionally introduce a recovery target $T_{\mathrm{rec}}$ (e.g., a \emph{get-up} prefix) and use a two-stage plan: first reach $T_{\mathrm{rec}}$, then replan to $T_{\mathrm{cmd}}$.

% % \emph{Safe halt and stationarity.}
% % During e-stop, we override the policy with a damping controller and wait until the system is stationary (e.g., CoM and joint velocities below thresholds) before attempting the recovery plan.

\subsubsection{Online Skill Scheduling}

The planner is invoked under four conditions: initialization, a user-commanded target change, approaching the end of the current reference, and safety-threshold events when $\mathrm{sim}$ crosses $A$ or $B$. 
On a trigger, the scheduler performs the RoA-based entry check, forms \texttt{candidates}, and invokes the selected planner with $T=T_{\mathrm{cmd}}$  
(\emph{for the NN planner only}, if a direct switch toward $T_{\mathrm{cmd}}$ is unsafe, we optionally use a two-stage call with a recovery target $T_{\mathrm{rec}}$ before returning to $T_{\mathrm{cmd}}$). 
With Graph Search, caching $V^{\*}$ and \texttt{NextHop} makes replanning a nearest-neighbor lookup plus pointer chasing; With NN, replanning is just the nearest-neighbor lookup. Both yield low online latency.

%% file: text/4_experiment.tex
\section{Experiment}
\begin{figure*}[t!]
\begin{center}
  \includegraphics[width=\textwidth]{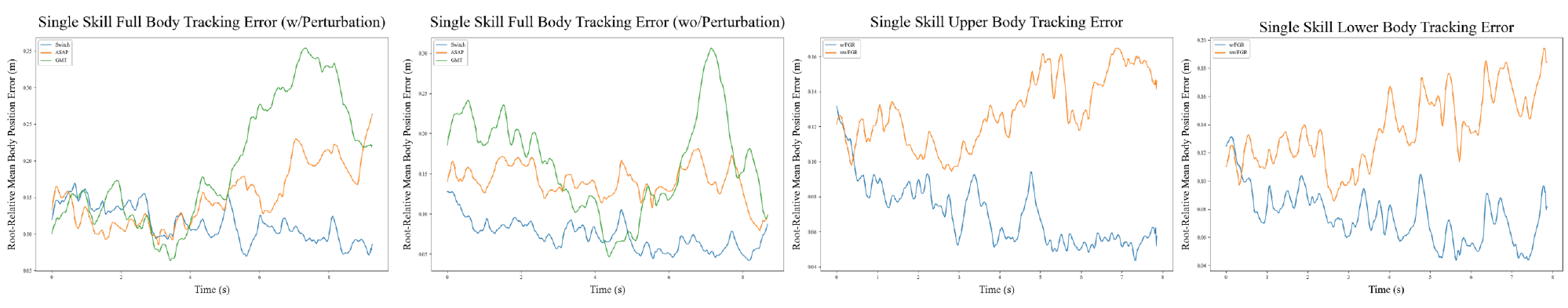}
\end{center}
  \caption{ Quantitative Evaluation of Skill Execution Performance. We analyzes the full-body tracking performance of the proposed \ours, alongside baselines ASAP and GMT, under both perturbed and unperturbed conditions. The results show that Switch maintains lower overall full-body tracking errors in single-skill execution experiments regardless of the presence of perturbation demonstraing its robotness to disturbance. Additionally, we also investigate the effectiveness of the Foot-Ground Contact Reward (FGR) on tracking performance. The results shows that it enhance full body motion with a more distinct improvement observed in the lower body motion tracking.  } 
  \label{fig: mot perf}
\end{figure*}

\begin{table*}[h!]
    \centering
    \caption{\textbf{Sim2Sim Comparison in MuJoCo}. We compare \ours with ablated versions and GMT across three task difficulty levels. \ours consistently achieves the highest Skill Switching Success Rate (SSR) and lowest motion errors, highlighting the effectiveness of each component. Here we use SG to represent the skill graph, B for the buffer state inpainting, and C for the foot-ground contact reward.}
    \vspace{3pt}
    \renewcommand{\arraystretch}{1.5}
    \begin{tabular}{lcccccccc}
    \toprule[1.0pt]
    \textbf{Method} & SSR ($\%$) $\uparrow$ & NR $\uparrow$ & $\text{E}_{\text{g-mpbpe}} \downarrow$ & $\text{E}_{\text{mpbpe}} \downarrow$ & $\text{E}_{\text{mpjpe}} \downarrow$ & $\text{E}_{\text{mpjve}} \downarrow$ &  $\text{E}_{\text{mpbve}} \downarrow$ &  $\text{E}_{\text{mpbae}}  \downarrow$  \\
    
    \midrule[0.6pt]
    \rowcolor[gray]{0.9} \multicolumn{9}{l}{\textbf{\textit{Easy}}} \\
    \midrule[0.6pt]
    GMT~\cite{chen2025gmt} 
    &30.00 & 2.636  &0.396 &0.392 &0.295 &1.420 & \textbf{0.391} & \textbf{5.651} \\
    Base & 2.00 & 4.520 & 0.120 & 0.118 &0.307 &\textbf{1.310} &0.398 &6.669 \\
    Base + SG & \textbf{100.0} & 5.802 &0.115 &0.121 &0.241 & 1.489 & 0.451 &7.883 \\
    Base + SG + B & \textbf{100.0} & 6.629 &0.087 &0.092 & \textbf{0.213} &1.472 & 0.446 & 8.089 \\
    Base + SG + B + C (\ours) & \textbf{100.0} & \textbf{7.029} & \textbf{0.075} & \textbf{0.078} & 0.220 & 1.483 & 0.454 &8.026  \\
    
    \midrule[0.6pt]
    \rowcolor[gray]{0.9} \multicolumn{9}{l}{\textbf{\textit{Medium}}} \\
    \midrule[0.6pt]
    GMT~\cite{chen2025gmt} 
    &10.00 &2.531  &0.491 &0.486 &0.304 &1.476 &0.397 & \textbf{5.681} \\
    Base & 2.00 & 4.505 &0.120 &0.120 &0.307 & \textbf{1.275} & \textbf{0.393} &6.560 \\
    Base + SG & \textbf{100.0} &5.175 &0.147 &0.151 &0.255 &1.464 &0.460 & 7.734 \\
    Base + SG + B & \textbf{100.0} & 6.128 & 0.103 & 0.106 & \textbf{0.214} &1.492 &0.459 & 8.304 \\
    Base + SG + B + C (\ours) & \textbf{100.0} & \textbf{6.616} &\textbf{0.090} &\textbf{0.098} &0.223 &1.484 &0.475 & 7.778 \\
    
    \midrule[0.6pt]
    \rowcolor[gray]{0.9} \multicolumn{9}{l}{\textbf{\textit{Hard}}} \\
    \midrule[0.6pt]
    GMT~\cite{chen2025gmt}
    &2.00 &2.877 &0.588 &0.591 &0.236 &\textbf{1.400} & \textbf{0.405} & \textbf{6.062} \\
    Base & 2.00 &3.570  &0.251 &0.253 &0.306 &1.630 &0.468 &6.416 \\
    Base + SG & \textbf{100.0} &5.091 &0.146 &0.150 &0.246 &1.531 & 0.469 & 8.129  \\
    Base + SG + B & \textbf{100.0} &6.015 &0.105 &0.105 & \textbf{0.213} & 1.446 &0.460 & 8.078 \\
    Base + SG + B + C (\ours) & \textbf{100.0} & \textbf{6.284}  & \textbf{0.098} & \textbf{0.093} &0.225 &1.458 &0.463 & 7.909 \\
    \bottomrule[1.0pt]
    \end{tabular}
    \label{table: main_table}
\end{table*}

% In this section, we present comprehensive experimental results evaluating skill transition and motion tracking performance on the Unitree G1, a 29 degree-of-freedom (DoF), 1.3m tall humanoid robot. We demonstrate that our method, \ours, is not only robust to agile skill transitions but also maintains high-fidelity motion tracking when performing individual skills. Additionally, we conduct an ablation study to analyze the key factors that influence \ours’s performance. 

% In this section, we evaluate \ours through comprehensive experiments, focusing on skill transition robustness and motion tracking accuracy.  We compare \ours against baseline methods, and conduct an ablation study to identify the key components driving its superior performance.

\subsection{Experiment Setup}

\subsubsection{Hardware Setup}
We conduct experiments on the Unitree G1 humanoid robot with 29 DoF. The learned policy runs onboard the robot using a Jetson Orin NX.

\subsubsection{Baseline} We compare \ours with its ablated versions and GMT~\cite{chen2025gmt}.
\begin{itemize}
\item{Base}: The basic RL policy trained on single skill data only using motion tracking method~\cite{he2025asap}.
\item{Base + SG}: Trained using motion tracking on Skill Graph-augmented trajectories.
% Trained with motion tracking method on Skill Graph augmented motion tracjactories without the buffer state inpainting and Foot-Ground Contact Reward (FGR).
\item{Base + SG + B}: Trained using motion tracking on Skill Graph-augmented trajectories with buffer states.
% Trained with motion tracking method on Skill Graph augmented motion tracjactories with buffer states but FGR is not involved.
\item{Base + SG + B + C (\ours)}: Trained using motion tracking on Skill Graph-augmented trajectories with buffer states and incorporating Foot-Ground Contact Reward (FGR).
% Trained with motion tracking method on Skill Graph augmented motion tracjactories with buffer states and FGR is involved.
\item{GMT}~\cite{chen2025gmt}: A state-of-the-art general tracking model.
% Pretrained state-of-the-art general tracking model.
\end{itemize}

\subsubsection{Evaluation Tasks}
We evaluate the policy’s tracking performance using a motion dataset that encompasses 4 distinct skills. Skill transitions are categorized into three difficulty levels, easy, medium, and hard, based on their number of transition to perform. Specifically, an "easy" transition refers to switching between skills once, a "medium" transition involves switching skills twice, and a "hard" transition refers switching skills three times. For each of these difficulty settings, the policies are trained in IsaacGym and subsequently evaluated in MuJoCo~\cite{todorov2012mujoco} over 50 trials each.

\subsubsection{Evaluation Metric}

We employ a set of metrics to assess both the skill transition effectiveness and motion imitation accuracy of the proposed method. First, the Skill Switching Success Rate (SSR) quantifies the percentage of successful target skill transitions when the policy is initialized from an arbitrary state different from the target skill; a transition is deemed unsuccessful if the average body position error relative to the root frame exceeds 0.5 meters at any point during imitation. Second, the Normalized Reward (NR) calculates the average reward per frame, formulated as $NR = \frac{1}{T} \sum_{t=0}^{T-1} \bar{r}_t$ where \(\bar{r}_t\) denotes the normalized reward at time \(t\) and \(T\) is the total number of frames processed. Additionally, we evaluate motion imitation fidelity via multiple tracking error metrics: Global Mean Per Body Position Error (\(E_{\text{g-mpbpe}}\), in meters), Root-Relative Mean Per Body Position Error (\(E_{\text{mpbpe}}\), in meters), Mean Per Joint Position Error (\(E_{\text{mpjpe}}\), in radians), Mean Per Joint Velocity Error (\(E_{\text{mpjve}}\), in radians per frame), Mean Per Body Velocity Error (\(E_{\text{mpbve}}\), in meters per frame), and Mean Per Body Acceleration Error (\(E_{\text{mpbae}}\), in meters per frame\(^2\)).

\subsection{Experimental Results}

\begin{figure}[t!] % [h!] suggests placing the figure "here" if possible
    \centering % Centers the figure horizontally
    \includegraphics[width=\linewidth]{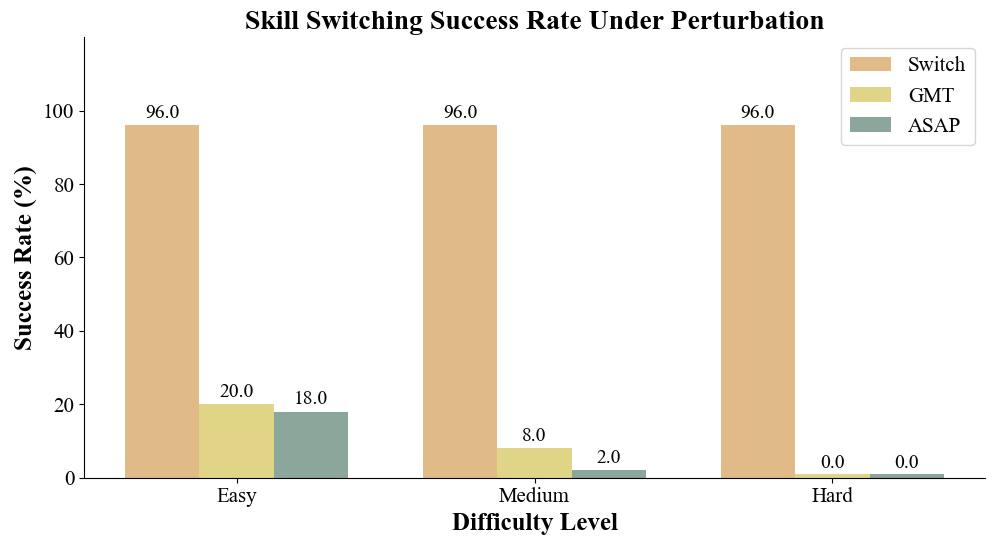} % Adjust width as needed
    % \caption{\textbf{Skill Transition Sucess Rate Under Pertubation}. To acess the robustness of the policy, we evaluate \ours by randomly pushing it with 25 N. We compare the skill transition rate with different methods using the same pertubation to show the stablility of switch.  }
    \caption{\textbf{Skill switching success rate under perturbation.} The perturbation is random 25 N pushes in simulation.}
    \label{fig:tsr_pert} % Optional: for referencing the figure
\end{figure}

\subsubsection{Performance in Skills Switching}
As summarized in Table~\ref{table: main_table}, \ours (denoted as Base + SG + B + C) achieves consistent and superior performance across all difficulty levels, outperforming all compared baselines. Regarding SSR, the Base model, trained solely on single-skill data stagnates at a mere 2.00\% across all levels, failing to execute even simple skill switching. The state-of-the-art general tracking model GMT~\cite{chen2025gmt} exhibits a drastic decline in SSR with increasing difficulty, dropping from 30.00\% (Easy) to 2.00\% (Hard), showing the limitation of general tracking model's reliance on pre-defined trajectories with feasible transition states in skill switching tasks. In contrast, models integrating the Skill Graph (SG) (Base + SG, Base + SG + B, \ours) attain a perfect 100\% SSR across all levels, proving the effectiveness of SG in enabling cross-skill transitions learning.

\begin{figure}[t!] % [h!] suggests placing the figure "here" if possible
    \centering % Centers the figure horizontally
    \includegraphics[width=\linewidth]{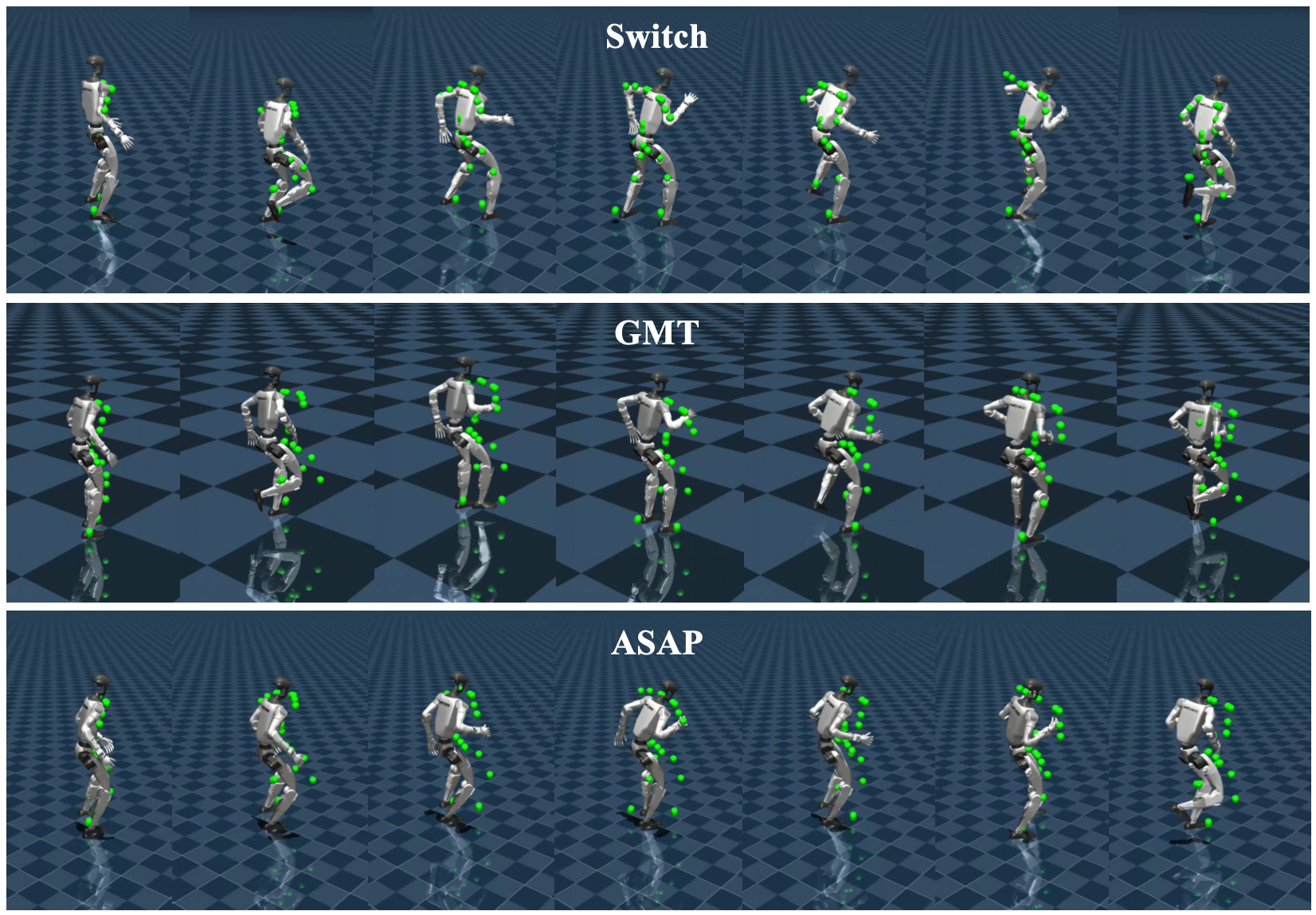} % Adjust width as needed
    \caption{Visual comparisons of skill execution (wo/Perturbation) across different methods in dancing which requires high-frequency foot-ground contact. \ours demonstrates more coordinated lower-body movement and attains better foot-ground interaction while ASAP~\cite{he2025asap} and GMT~\cite{chen2025gmt} exhibit conservative and jerky lower body motions, particularly when agile movements are required.}
    \label{fig:simvis} % Optional: for referencing the figure
\end{figure}

\subsubsection{Performance in Skills Execution}
In terms of performance on skill execution, \ours maintains the lowest errors in key metrics across most difficulty levels. From Table~\ref{table: main_table}, its Global Mean Per Body Position Error  is 0.075m (Easy), 0.090m (Medium), and 0.098m (Hard), substantially lower than GMT’s 0.396m, 0.491m, and 0.588m, and also outperforming Base + SG + B (0.087m, 0.103m, 0.105m). Similarly, Switch achieves the smallest Root-Relative Mean Per Body Position Error  (0.078m, 0.098m, 0.093m), reflecting precise whole-body motion alignment with reference trajectories. The improvement is particularly pronounced in lower-body motion as shown in Figure~\ref{fig:simvis}, which demonstrates Switch’s more coordinated lower-body movements compared to GMT~\cite{chen2025gmt} and ASAP~\cite{he2025asap}. Critically, even in the Hard difficulty level (involving three consecutive skill switches) as shown in Figure~\ref{fig: mot perf}, Switch retains low tracking errors, demonstrating its robustness to perturbation and escalating task complexity.

\subsubsection{Effectiveness of Foot-Ground Contact Reward}

To investigate the impact of the Foot-Ground Contact Reward (FGR) on motion tracking, we compared the upper and lower body motion tracking performance for individual skills by assessing our policy trained both with and without FGR. From Figure~\ref{fig: mot perf}, we found that when FGR is incorporated, the tracking performance improves, with the lower body exhibiting a more pronounced enhancement . This is attributed to the fact that the foot-ground contact reward directly strengthens the interaction between the robot’s lower body and the ground, refining the precision of lower-body motion. Figure~\ref{fig:simvis} presents the qualitative comparsion between \ours and other motion tracking baseline. It can be observed that \ours achieves better foot-ground contact compared to GMT and ASAP on agile skill execution such as dancing, and demonstrates stable, high-fidelity motion imitation.

%% file: text/5_conclusion.tex
\section{Conclusion}

In this paper, we propose \ours , which addresses the critical challenge of multi-skill switching in humanoid robots by integrating a Skill Graph for data augmentation, a reinforced whole-body tracking controller with buffer states and foot-ground contact optimization, and an online scheduler for real-time planning. Unlike baselines that struggle with increasing task difficulty or perturbations, Switch achieves 100\% SSR across all skill switching levels, maintains low tracking errors, and autonomously recovers from disturbances, demonstrating its ability to enable agile, seamless skill execution and laying a foundation for practical deployment of humanoid robots in dynamic real-world scenarios.
